\pdfoutput=1
\documentclass[a4paper]{article}

\usepackage{INTERSPEECH2020}
\usepackage{multirow,xcolor}
\usepackage{amsmath}

\title{Modeling ASR Ambiguity for Neural Dialogue State Tracking}
\name{Vaishali Pal$^{1,3}$, Fabien Guillot$^1$, Manish Shrivastava$^3$, Jean-Michel Renders$^1$, Laurent Besacier$^{1,2}$}
\address{
  $^1$Naver Labs Europe\\
  $^2$ LIG - Universit\'e Grenoble Alpes, France\\
  $^3$International Institute of Information Technology, Hyderabad, India}
\email{vaishali.pal@research.iiit.ac.in, jean-michel.renders@naverlabs.com, laurent.besacier@univ-grenoble-alpes.fr}

\begin{document}

\maketitle
\begin{abstract}
  Spoken dialogue systems typically use one or several (top-N) ASR sequence(s) for inferring the semantic meaning and tracking the state of the dialogue However, ASR graphs, such as confusion networks (confnets), provide a compact representation of a richer hypothesis space than a top-N ASR list. In this paper, we study the benefits of using confusion networks with a  neural dialogue state tracker (DST). We encode the 2-dimensional confnet into a 1-dimensional sequence of embeddings using a confusion network encoder which can be used with any DST system. Our confnet encoder is plugged into  the  `Global-locally Self-Attentive Dialogue State Tacker' (GLAD) model for DST and obtains significant improvements in both accuracy and inference time compared to using top-N ASR hypotheses.
\end{abstract}

\section{Introduction}
\label{sec:intro}

Spoken task-oriented dialogue systems guide the user to complete a certain task through speech interaction. While such speech systems generally include an explicit automatic speech recognition (ASR) module (\textit{cascade} approach), they now tend to be replaced by \textit{end-to-end} approaches where the systems take speech as input and directly produce a decision from it. Examples include end-to-end architectures for spoken language understanding (SLU) proposed recently \cite{DBLP:journals/corr/abs-1802-08395,DBLP:journals/corr/abs-1906-07601,DBLP:journals/corr/abs-1904-03670,dinarelli2020data}.
However, those end-to-end models currently lead to equivalent but no better performance compared to \textit{cascade} approaches based on ASR  (see for instance \cite{DBLP:journals/corr/abs-1802-08395,DBLP:journals/corr/abs-1906-07601}). Besides, in some specific use cases, it may be preferable to deploy modular systems (instead of a monolithic one) for which only one component (ASR, SLU, dialog state tracker) can be modified at a time.

\begin{figure}[h!]
 \centering
 \includegraphics[scale=0.35]{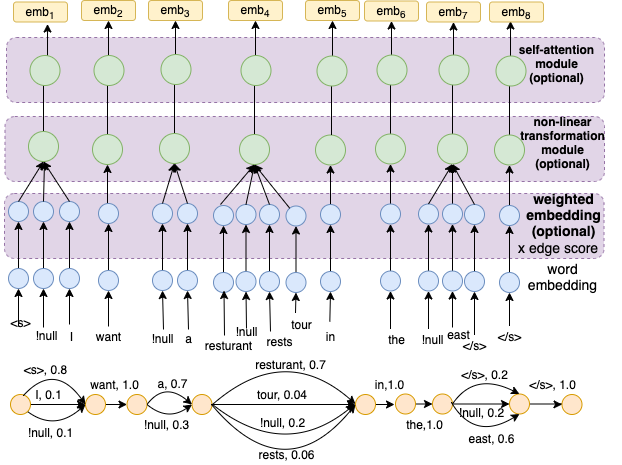}
 \caption{Word Confusion Network Encoder Variations}
 \label{fig:confnet_encoder}
 \end{figure}

This article is positioned in this latter context and studies how to take better account of ASR ambiguity in voice-based dialog state tracking systems. Most recent work on spoken dialogue systems uses one or several (top-N) ASR sequence(s) to track the dialogue state and infer user needs. However, ASR lattices provide a richer hypothesis space than the top-N hypotheses. More precisely, we revisit the use of word confusion networks (simply denoted as confnets) \cite{DBLP:journals/csl/ManguBS00}, derived from ASR lattices, as a compact and efficient representation of ASR output. To encode such graphical representations with existing state-of-the-art dialogue state trackers (DST),
%allows us to increasing their scope. 
%Inspired from recent work on encoding ASR lattices and confusion network, 
we introduce a generic neural confusion network encoder (see Figure \ref{fig:confnet_encoder}) which can be used as a plug-in to any dialogue state tracker and achieves better results than using a list of top-N ASR hypotheses.

Our research contributions are the following: %\laurent{[LB: to be updated according to our findings]}
\begin{itemize}
  \item we introduce a system which encodes the confusion network into a representation that can be used as a plug-in to any state-of-the-art dialogue state tracker,
  \item we propose and experiment several variants to encode ASR confnets,
  \item we introduce mechanisms to leverage both ASR confusion network and true transcripts while training the DST system,
  \item we explore DST using a richer hypothesis space in the form of a confusion network and study whether it leads to better  performance / computation trade-off  compared to using a list of top-N ASR hypotheses.
\end{itemize}

\section{Related Work}
\textbf{Dialog state tracking.} Recent pieces of work on dialogue state trackers \cite{zhong-etal-2018-global,DBLP:journals/corr/abs-1810-09587,DBLP:journals/corr/MrksicSWTY16} infer the state of the dialogue from conversational history and current user utterance. These systems assume a text-based user utterance and accumulate the user goal across multiple user turns in the dialogue. \cite{zhong-etal-2018-global} generalizes on rare slot-value pairs by using global modules which share parameters and local modules to learn slot specific feature representations.  \cite{DBLP:journals/corr/abs-1810-09587} achieves state-of-the-art performance on DSTC-2 dataset \cite{henderson-etal-2014-second} with a universal state tracker which generates fixed-length representation for each slot and compares the distance between the representation and value vectors to make predictions. \cite{Chen2020SchemaGuidedMD} predicts dialogue states
from utterances and schema graphs containing
slot relations in edges to achieve state-of-the-art result of MultiWoz 2.0 \cite{budzianowski-etal-2018-multiwoz} and MultiWoz 2.1 datasets \cite{DBLP:journals/corr/abs-1907-01669}.

\textbf{Using ASR graphs with neural models.} 
Word lattices from ASR were used by \cite{Ladhak2016LatticeRnnRN} for intent classification in SLU with RNNs. Inspired from \cite{Ladhak2016LatticeRnnRN}, \cite{jagfeld-vu-2017-encoding} proposed to use word confusion networks for DST. However, there are several differences with what we propose in our paper: (1) they only use average pooling to aggregate the hidden GRU states corresponding to the alternative word hypotheses whereas we introduce several variants to pool alternative words (see section \ref{sec:approach}), (2) our word confusion network encoder can be plugged into any neural architecture for dialogue state tracking, as it basically amounts to have a first layer that transforms a 2D-data structure (confnets) into a 1D-sequence of embeddings, while \cite{jagfeld-vu-2017-encoding} keeps the 2D-data structure in the hidden layers and, consequently, is limited to simple RNNs such as GRU and (bi-)LSTM, and (3) they experiment with a simpler RNN-based dialog state tracker while we plug our confnet encoder into  the more efficient 'Global-locally Self-Attentive Dialogue State Tacker' (GLAD) model of \cite{zhong-etal-2018-global}. 
Finally, our confnet encoder is most similar to \cite{masamura-et-al-2018} but they used ASR confnets  for classification of user intent, question-type and named-entities
%Our work is inspired by \cite{masamura-et-al-2018} and encodes a confusion network into a representation space which can be used in any state-of-the-art dialogue state trackers. While \cite{masamura-et-al-2018} classifies user intent with the confusion network encoder,
while we apply our encoder to a DST task (we also propose several variants over \cite{masamura-et-al-2018}).
\section{Word Confusion Network for DST}
\label{sec:approach}

\subsection{Confusion Network Encoder}
\label{ssec:encoder}
Inspired from \cite{masamura-et-al-2018}, we use a word confusion network encoder to transform the graph to a representation space which can be used with any dialogue state tracker. The multiple aligned ASR hypotheses, represented as parallel arcs at each position of the confusion network, are treated as a set of weighted arcs.
%over which a self-attention layer is applied to emphasize more relevant words in aligned word space. The final embedding representation at time/position $t$ can be interpreted as an attended word embedding over the aligned word hypotheses at time/position $t$. 
More formally, a confnet $C$ is a sequence of parallel weighted arcs, noted as
$C=[(<w^1_1,\pi^1_1>,<w^2_1,\pi^2_1>,...,<w^{n_1}_1,\pi^{n_1}_1>),\ldots, (<w^1_m,\pi^1_m>,<w^2_m,\pi^2_m>,...,<w^{n_m}_m,\pi^{n_m}_m>)]$,
%$C=[(<w^1_1,\pi^1_1>,<w^1_2,\pi^1_2>,...,<w^1_{n_1},\pi^1_{n_1}>),\ldots, (<w^m_1,\pi^m_1>,<w^m_2,\pi^m_2>,...,<w^m_{n_m},\pi^m_{n_m}>)]$,
where $w^j_t$ is the $j^{th}$ arc (token) at time/position $t$, and $\pi^j_t$ its associated confidence weight. We propose several variants to formulate the embedding representation for a set $C_t$ of parallel arcs at position $t$ in the confnet $C$.

%can be formulated in multiple ways. 
The simplest method to encode the confusion network is as a sequence of weighted-sums of the word embeddings weighed with the ASR confidence scores:

\begin{eqnarray}
    p_t^i &=& \pi_t^i Embedding(w_t^i) \\
    e_{v1}(C_t) &=& \sum_i p_t^i
\end{eqnarray}
In increasing complexity, the second variant to encode the parallel arcs is to apply a weighted sum of non-linear transformations over word embeddings:
\begin{eqnarray}
    r_t^i &=& \pi_t^i\tanh( W_1 Embedding(w
    _t^i)) \\
    e_{v2}(C_t) &=& \sum_i r_t^i
\end{eqnarray}

The third variation is to formulate the encoding with a self-attention mechanism similar to that described in  \cite{masamura-et-al-2018}. In this case, $\pi_t^i$ (ASR) weights can be ignored as the model will use self-attention  to weight the parallel arcs. The self-attention weights $\alpha_t^i$ are learnt during training:
 \begin{eqnarray}
 q_t^i &=& \tanh( W_1 Embedding(w_t^i))\\
 \alpha_t^i &=& \frac{\exp(w_2^T q_t^i)} {\sum_j \exp(w_2^T q_t^j)}\\
 e_{v3}(C_t) &=& \sum_i \alpha_t^i q_t^i
 \end{eqnarray}
$Embedding(w_t^i)$ is the embedding representation of word $w_t^i$.
%which is the concatenated GloVe and Kazuma character embeddings. 
The final variation is to use the self-attention mechanism exactly as described in  \cite{masamura-et-al-2018}. The ASR weights $\pi^i_t$ are used as an additional feature to weigh the word embeddings of each parallel arc:
\begin{eqnarray}
\label{eqs}
    \bar{q_t^i} &=& \tanh( W_1 p_t^i)\\
    \bar{\alpha_t^i} &=& \frac{\exp(w_2^T \bar{q_t^i})} {\sum_j \exp(w_2^T \bar{q_t^j)}}\\
    e_{v4}(C_t) &=& \sum_i \bar{\alpha_t^i} \bar{q_t^i}
\end{eqnarray}  
$e_{vx}$ denotes the 4 variations of the standard trainable embedding layer for word/token $w_t^i$; the matrix $W_1$ and the vector $w_2$ are  trainable parameters of our model. Note that the training of these parameters is done jointly with the main task (see next subsection). 

\subsection{Dialogue State Tracking with Confnet}
The dialogue state is a representation of the user goal at any time in the dialogue.  A dialogue state tracker (DST) accumulates  evidence as the dialogue progresses at each user turn and updates the state to reflect the changing user goals. The user-goal is captured by the tracker as a distribution of slot-value pairs. Each user utterance can be either in textual or spoken form. Conventionally,  DST uses top-N list of ASR hypotheses of the spoken user utterances to track the user needs. However, graph based representation such as ASR lattices and confusion network provides a richer hypothesis space in compact form.

Our confusion network encoder can be used as a plug-in to any state-of-the-art DST system. We have used the  'Global-locally Self-Attentive Dialogue State Tacker' (GLAD) model \cite{zhong-etal-2018-global} with our confusion network encoder. GLAD addresses the issue of rare slot-value pairs which were not explicitly handled by previous DST models. The GLAD encoder module is a global-local self-attentive encoder which separately encodes the transcript/ASR hypothesis, system actions from previous turns and slot-value under consideration. We extend GLAD by replacing the user utterance representation, namely a sequence of trainable token embeddings, by the confnet embedding sequence. Remind that a confnet is also encoded as a 1 dimensional sequence 
%embedding encoded by the confnet encoder is a sequence 
of embeddings that corresponds to each time/position in the confnet.  This enables  GLAD architecture to use graph-based inputs instead of (or even in addition to)  token sequence inputs.

\section{Model Training Strategies}

At training time, both clean transcript and ASR graph are available. It is therefore tempting to use these two pieces of information to facilitate  model training while trying to make it  robust to ASR errors in the meantime. We propose two radically different strategies to take into account clean transcript \textit{and} ASR graph at training time.
\begin{figure}[h!]
 \centering
 \includegraphics[scale=0.35]{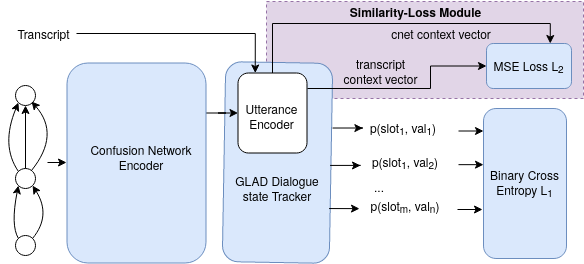}
 \caption{Confusion Network Encoder used with DST system (GLAD). Similarity-loss module can be used for augmenting the transcript data to jointly-train the system.}
 \label{fig:manual}
\end{figure} 

 \begin{figure*}[h!]
 \centering
 \includegraphics[scale=0.5]{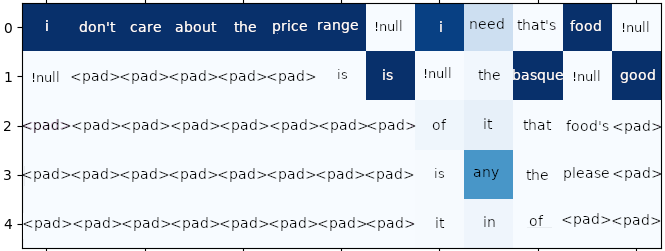}
 \caption{Attention Weights of the confusion network encoder (variant $e_{v_4}$) for the confnet with transcript `i don't care about the price range i need basque food'. Dark colors represent high values and light colors low values. The columns represent subsequent parallel arcs. The parallel arcs are sorted from highest to lowest scored hypothesis. The 1st row is the best-pass through the network. The highest attention-weights are for the words `i don't care about the price range is i any basque food good'}
 \label{fig}
\end{figure*} 

\subsection{Data Augmentation}
The confusion network contains noisy hypotheses with lower ASR confidence scores which makes training hard. Augmenting the confnet dataset with the clean transcript should help the system to converge faster and better. 
%as shown in the figure \ref{fig:jg_inference_time}. 
We encode transcript in the form of a confnet with a single arc between nodes. At training time we merge both noisy ASR and clean (single arc) confnet datasets. Consequently, we use each dialog twice at training.

\subsection{Adding a Similarity Loss to Train Confnet Embeddings: Jointly-Trained Model}
\label{similarity}

We want the confnet representation to be very close to the one of the clean transcript in the embedding space. To enforce this, we add a similarity loss to the loss corresponding to our main task (binary classification for each $<$slot,value$>$ pair) - see figure \ref{fig:manual}.

Binary-cross-entropy is used as classification loss, denoted as $L_{1}$, and squared euclidean distance between clean and noisy transcripts is used as similarity loss ($L_2$). The binary cross-entropy loss can be modeled as shown in equation \ref{eq:loss} where $p(slot_m, val_n)$ represents the prediction for the slot $m$ and value $n$ and $y$ is the binary ground-truth. Let $Embedding_{CN}$ be the transformation realised by the confnet encoder and $Embedding_T$ be the standard embedding layer on word tokens; $f$ be the transformation function of the GLAD encoder that takes as input the sequence of one dimensional embeddings (either standard word token embeddings or confnet embeddings) and outputs a global-local context vector. The final loss, $L$, of the model is defined as a linear combination of the cross-entropy loss, $L_1$, and the similarity loss, $L_2$ as shown in equation \ref{eq:loss}. We choose a $\lambda$ value of 0.5 for our experiments. 
\begin{equation}
\begin{aligned}
     L &= L_1 + \lambda L_2
         \label{eq:loss}
\end{aligned}
\end{equation}
 \begin{eqnarray}
  L_1 &=& -(y\log(P) + (1-y)\log(1-P)) \\
  P &=& p(slot_{m}, val_{n})
 \end{eqnarray}
\begin{equation}
    \begin{split}
        L_2 = \|f(Embedding_T(transcript)) \\ - f(Embedding_{CN}(confnet))\|^2_2 
        \end{split}
\end{equation}
 Note that, while $L_2$ alone has a trivial solution (namely, projecting everything to a null vector), the combination with the loss associated to the primary task renders this trivial solution non-optimal. The model trained jointly using the similarity loss $L_2$ is called \emph{JCnet}.
 
\section{DSTC-2 Dataset}
We evaluate our system on the standard Dialogue State Tracking Challenge 2 (DSTC-2) dataset \cite{henderson-etal-2014-second}.\footnote{http://camdial.org/~mh521/dstc/} DSTC is a research challenge focused on improving the state-of-the art in tracking the state of spoken dialog systems. DSTC 2 contains dialogue in the restaurant information domain where the states may change through the dialogue.
%The dataset contains dialogues from the restaurant information domain which changing user goals. 
There are 1612, 506 and 1117 dialogues in the training,  development and test sets respectively. The dialog state is captured by  \textit{informable} slots (slots which the user can provide a value for, to use as a constraint on their search) and  \textit{non informable} (unconstrained) slots.
%The dialogue state is captured by informable slots.
%The dialogue state is captured by informable slots which constraints the possible values the slot take have and non-informable slots which is unconstrained. \laurent{[LB: the previous sentence should be rephrased...]} 
All the slots in the dataset are requestable as the user can request the value of any slot. For example, for a given restaurant, the user can request the value of the phone number or price-range slot. At each turn of the dialogue, the user's goal may change, new information maybe requested or provided to the system by the user. Thus, a turn is represented as a dialogue state comprising of the triplet (user-goal, turn-request, turn-inform). DSTC-2 provides representation of the user speech utterance in the form of top-10 ASR hypotheses and word confusion networks.

% The confusion network is a list of parallel arcs (sausages) given by the ASR which contains the following attributes:

% \begin{itemize}
%     \item \hl{\textbf{start}: time in seconds (float) into the user turn this set of parallel arcs (aligned hypotheses) starts.}
%     \item \hl{\textbf{end}: time in seconds (float) into the user turn this set of parallel arcs (aligned hypotheses) ends.}
%     \item \hl{\textbf{arcs}:  a list of aligned hypotheses in this sausage:}
%     \begin{itemize}
%         \item \hl{\textbf{word}: the hypothesis this arc corresponds to. Includes a special "!null" for an empty arc.}
%         \item \hl{\textbf{score}:  the log probability of this arc.}
%     \end{itemize}
% \end{itemize}

We followed a similar pre-processing pipeline of the confusion networks as mentioned in \cite{jagfeld-vu-2017-encoding}, i.e, we removed the interjections (um, ah, etc) and pruned arcs with an ASR confidence score less than 0.001 to reduce the size of the network, thus increasing efficiency without compromising the accuracy of models. To obtain a fair and consistent performance comparison between models based on ASR-top-N hypotheses and on confusion network, the top-N list of ASR hypotheses was chosen as the N best paths extracted from the confusion network.
%these top-10 best paths.
%are further processed to remove the occurrences of the $!null$ token, which represents an empty arc with no semantic value. 

\section{Experiments and Results}

% \begin{table*}[!t]
%     \centering
%     \begin{tabular}{|c|c|c|c|c|c|c|c|c|c|c|c|c|c|c|c|c|c|c|c|c|c|c|}
%     \hline
%   N &
%   \multicolumn{8}{c|}{1}
%   \multicolumn{8}{c|}{5} &
%     \multicolumn{8}{c|}{9} \\
%     \cline{2-19}
%       &  \multicolumn{2}{c|}{JG} &
%     \multicolumn{2}{c|}{TI} &
%     \multicolumn{2}{c|}{TR}
%      &  \multicolumn{2}{c|}{JG} &
%     \multicolumn{2}{c|}{TI} &
%     \multicolumn{2}{c|}{TR}
%      &  \multicolumn{2}{c|}{JG} &
%     \multicolumn{2}{c|}{TI} &
%     \multicolumn{2}{c|}{TR}
%     \\
%     \cline{2-19} 
%     & $\mu$ & $\sigma$  & $\mu$ & $\sigma$ & $\mu$ & $\sigma$ & $\mu$ & $\sigma$  & $\mu$ & $\sigma$ & $\mu$ & $\sigma$
%     & $\mu$ & $\sigma$ & $\mu$ & $\sigma$
%     & $\mu$ & $\sigma$ 
%     \\
%     \hline
%     ASR-N  & & & & & & & & & & & & & & & & & & \\
%     \hline
%     Aug ASR-N & 68.46 & 0.0017 & 83.26 & 0.0012 & 96.68 & 0.0003 & & & & & &  & & & & & & \\
%     \hline
%     \end{tabular}
%     \caption{Scores for model trained on ASR list of hypothesis.}
%     \label{tab:ASR_scores}
% \end{table*}

%   \begin{figure}[h!]
%  \centering
%  \includegraphics[scale=0.6]{jg_time.png}
%  \caption{Joint-Goal vs. Inference Time: Each point is annotated with number of hypothesis used for inference.}
%  \label{fig:time}
%  \end{figure}

\begin{table}[!b]
    \centering
        \caption{Scores for baseline model trained on augmented ASR-N list of hypothesis. Each cell contains the mean accuracy $\mu$(0-1) and standard error $\sigma$(0-1) in the format $\mu$($\sigma$) for 4 runs of each setting}
    \label{tab:ASR_scores}
    \begin{tabular}{|c|c|c|c|c|c|c|}
    \hline
   List-Size &  Joint-Goal & Turn-Inform & Turn-Request
    \\
    \hline
    ASR-1 & 0.6846(.0017) & 0.8326(.0012) & 0.9668(.0003) \\
    ASR-5 & 0.6980(.0075) & 0.8375(.0050) & 0.9680(.0010) \\
    ASR-9 & 0.6942(.0077) & 0.8395(.0055) & 0.9680(.0073) \\
    \hline
    \end{tabular}
\end{table}

\begin{table}[!b]
    \centering
        \caption{Scores for Jointly-trained (Similarity Loss) model on augmented confnet dataset. Each cell contains the mean accuracy  $\mu$ (0-1) and standard error $\sigma$ (0-1) in the format $\mu$($\sigma$) for 4 runs of each setting }
    \begin{tabular}{|c|c|c|c|c|}
    \hline
   Arc-Size & Joint-Goal  & Turn-Inform & Turn-Request
    \\
    \hline
    JCnet-1 & 0.6883(.0027) & 0.8361(.0021)& 0.9672(.0001) \\
    JCnet-5 & 0.7088(.0021) & 0.8403(.0015) & 0.9773(.0003)  \\
    JCnet-9 & 0.7063(.0049) & 0.8461(.0048) & 0.9700(.0005)\\
    \hline
    \end{tabular}
    \label{tab:simlilarity_scores}
\end{table}

\begin{table*}
  \caption{Results on confusion network model encoding variants with different numbers of parallel arcs (5 or 9). With or w/o confnet augmentation. Each cell contains the mean accuracy $\mu$(0-1) and standard error $\sigma$(0-1) of the 3 metrics on 10 runs of each setting in the format $\mu$($\sigma$)}
  \begin{tabular}{|c|c|c|c||c|c|c|}
    \hline
    Accuracy [$\mu$ ($\sigma$)] & Joint-Goal    & Turn-Inform   & Turn-Request & Joint-Goal & Turn-Inform & Turn-Request \\
    \hline
    \textbf{Parallel Arcs: 5} & \multicolumn{3}{c||}{\textbf{Confnet Encoding Variation 1}: $\boldsymbol{e_{v1} = \sum_i p_t^i}$} &
      \multicolumn{3}{c|}{\textbf{Confnet Encoding Variation 2}: $\boldsymbol{e_{v2} = \sum_i r_t^i}$}\\
   \hline 
    Non-aug CNET-N & 0.7019(0.0026) & 0.8350(0.003) & 0.9690(0.0004) & 0.7008(0.0027) & 0.8389(0.002) & 0.9677(0.0003)  \\
    \hline
    Aug CNET-N & 0.7121(0.0019) &  0.8470(0.0011) & 0.9688 (0.0004) & 0.7115(0.0019) & 0.8465(0.0011) & 0.9686(0.0005) \\
    \hline
     & \multicolumn{3}{c||}{\textbf{Confnet Encoding Variation 3}: $\boldsymbol{e_{v3} = \sum_i \alpha_t^i q_t^i}$} &
      \multicolumn{3}{c|}{\textbf{Confnet Encoding Variation 4}: $\boldsymbol{e_{v4}\sum_i\bar{\alpha_t^i}\overline{\bar{q_t^i}}}$}\\
   \hline
    Non-aug CNET-N & 0.6912(0.0057) & 0.8302(0.0037) & 0.9645(0.0005) & 0.6975(0.0016) &  0.8361(0.0013) & 0.9686(0.0004)  \\
    \hline
    Aug CNET-N & 0.6925(0.0032) &  0.8372(0.0008) & 0.9650 (0.0006) & 0.7056(0.0023) & 0.8413(0.0015) & 0.9689(0.0007)\\
    \hline
    \hline
    \textbf{Parallel Arcs: 9} & \multicolumn{3}{c||}{\textbf{ Confnet Encoding Variation 1}: $\boldsymbol{e_{v1} = \sum_i p_t^i}$} &
      \multicolumn{3}{c|}{\textbf{Confnet Encoding Variation 2}: $\boldsymbol{e_{v2} = \sum_i r_t^i}$}\\
   \hline 
    Non-aug CNET-N & 0.7044(0.0020) & 0.8395(0.0009) & 0.9693(0.0003) & 0.6997(0.0043) &  0.8349(0.0038) & 0.9684(0.0003)  \\
    \hline
    Aug CNET-N & 0.7094(0.0017) &  0.8446(0.0012) & 0.9694 (0.0004) & 0.7071(0.0095) & 0.8413(0.0057) & 0.9680(0.0004)\\
    \hline
     & \multicolumn{3}{c||}{\textbf{Confnet Encoding Variation 3}: $\boldsymbol{e_{v3} = \sum_i \alpha_t^i q_t^i}$} &
      \multicolumn{3}{c|}{\textbf{Confnet Encoding Variation 4}: $\boldsymbol{e_{v4}=\sum_i\overline{\bar{\alpha_t^i}}\bar{q_t^i}}$}\\
   \hline
    Non-aug CNET-N & 0.6938(0.0033) & 0.8325(0.0022) & 0.9650(0.0003) & 0.6969(0.0023) & 0.8382(0.0011) & 0.9680(0.0006)  \\
    \hline
    Aug CNET-N & 0.6930(0.0021) & 0.8367(0.0007) & 0.9648(0.0006)  & 0.7045(0.0023) & 0.8416(0.0011) & 0.9681(0.0008)\\
    \hline
  \end{tabular}
  \label{tab:joint_goal_confnet}
\end{table*}

Our baseline is the model trained on augmented dataset composed of ASR-N hypotheses and transcripts, where the final prediction is the weighted sum of prediction probabilities from each ASR hypothesis (\textit{Augmented ASR-N}, see table \ref{tab:ASR_scores}). To demonstrate that a richer hypothesis space (confusion network) helps in improving accuracy, we trained 3 separate models on confusion networks with non-augmented (\textit{Non-augmented Cnet-N}), augmented (\textit{Augmented Cnet-N}) dataset
and jointly-trained model (\textit{JCnet-N}). In these models, we restricted the number of arcs per token by keeping only the ones with the top-$N$ weights ($N \in [5,9]$).  
The augmented dataset is composed of transcripts modeled as a graph with one arc between nodes, and ASR confusion networks. In addition to the above data augmentation techniques, we also evaluate the impact of using a similarity loss to train the confnet embeddings as introduced in section \ref{similarity} (\textit{JCnet-N}). We  use a learning rate of $0.01$, a batch size of 50, a dropout of 0.2 and a $\lambda$ value of 0.5 to train our models. We concatenate pre-trained word embeddings (GloVe) \cite{Pennington14glove:global} and Kazuma character embeddings \cite{hashimoto-etal-2017-joint} to encode words. The embedding layer is frozen and not updated during training.

We train each experimental setting multiple times with different seeds and report the mean accuracy ($\mu$) and standard error ($\sigma$) of the joint-goal, turn-inform and turn-request for the ASR-N models in table \ref{tab:ASR_scores}, \emph{JCnet-N} models in table \ref{tab:simlilarity_scores} and the variants of confusion network encoder models in table \ref{tab:joint_goal_confnet}. Our comments will mostly focus on the joint-goal accuracy metric which is the most important. We use 10 runs to calculate $\mu$ and $\sigma$ for  \textit{Non-augmented Cnet-N} and \textit{Augmented Cnet-N} models and 4 runs to calculate $\mu$ and $\sigma$ for \textit{Augmented ASR-N} and \textit{JCnet-N} models. As illustrated in table \ref{tab:joint_goal_confnet}, the joint-goal accuracy of the models trained on \textit{augmented} dataset (leveraging clean transcripts at training time) performs better than those trained on \textit{non-augmented} one. These models also outperform the ASR-N baseline. The best variants seem to be $e_{v_1}$ and $e_{v_2}$ which do not use attention, both outperforming $e_{v_4}$ \cite{masamura-et-al-2018}. However, attention models like $e_{v_4}$ interestingly learn to assign higher weights to relevant words with lower ASR score over the top-1 hypothesis as shown in the figure \ref{fig}. Content words, such as `basque' which aids in classifier discriminability, is chosen by the attention-weights over function words such as `that's' inspite of the later being the top hypothesis. Furthermore, the performance of jointly-trained (similarity loss) confnet models in table \ref{tab:simlilarity_scores} perform similar to the data augmented confnet models in table \ref{tab:joint_goal_confnet}. Augmented confusion network with parallel arc size 5 (variant $e_{v_2}$) outperforms jointly-trained model by a margin of $1\%$ in joint-goal accuracy whereas both models have similar performance with parallel arc size 9.
%shows similar results in both the settings.

The confusion network models lead to significant inference time gains over those trained on the list of ASR hypothesis. ASR-N models aggregate the predictions over each hypothesis to formulate the final prediction, resulting in a time complexity of $O(NM)$ where $N$ is the number of ASR hypothesis and $M$ is the size of the neural network. The confusion network models eliminate the additional time complexity introduced by the ASR hypothesis size without compromising on the rich hypothesis space. Thus, all variations of the confusion network models have an inference time complexity of $O(M)$. Inference time for all the varying confnet models (augmented, non-augmented, and joint goal) is on an average $0.82 sec$ per batch over a batch-size of 50. The average inference per batch for the ASR-N models progressively increases from $0.57sec$ for ASR-1 to a maximum of $7 sec$ for ASR-10.

 \section{Conclusion}
In this paper, we have demonstrated that exploiting the rich hypothesis space of a confusion network instead of being limited to the top-N ASR hypotheses for DST leads to performance gain in time and accuracy. The time gain is significant if we want to incorporate a larger set of alternative ASR hypotheses. Moreover, we explore variations of designing the initial embedding layer transforming the confnet as a one-dimensional sequence of position-wise embeddings which can be plugged into numerous state-of-the-art text-based DST systems.

\bibliographystyle{IEEEtran}

\bibliography{template}

\begin{thebibliography}{10}
\providecommand{\url}[1]{#1}
\csname url@samestyle\endcsname
\providecommand{\newblock}{\relax}
\providecommand{\bibinfo}[2]{#2}
\providecommand{\BIBentrySTDinterwordspacing}{\spaceskip=0pt\relax}
\providecommand{\BIBentryALTinterwordstretchfactor}{4}
\providecommand{\BIBentryALTinterwordspacing}{\spaceskip=\fontdimen2\font plus
\BIBentryALTinterwordstretchfactor\fontdimen3\font minus
  \fontdimen4\font\relax}
\providecommand{\BIBforeignlanguage}[2]{{%
\expandafter\ifx\csname l@#1\endcsname\relax
\typeout{** WARNING: IEEEtran.bst: No hyphenation pattern has been}%
\typeout{** loaded for the language `#1'. Using the pattern for}%
\typeout{** the default language instead.}%
\else
\language=\csname l@#1\endcsname
\fi
#2}}
\providecommand{\BIBdecl}{\relax}
\BIBdecl

\bibitem{DBLP:journals/corr/abs-1802-08395}
\BIBentryALTinterwordspacing
D.~Serdyuk, Y.~Wang, C.~Fuegen, A.~Kumar, B.~Liu, and Y.~Bengio, ``Towards
  end-to-end spoken language understanding,'' \emph{CoRR}, vol. abs/1802.08395,
  2018. [Online]. Available: \url{http://arxiv.org/abs/1802.08395}
\BIBentrySTDinterwordspacing

\bibitem{DBLP:journals/corr/abs-1906-07601}
\BIBentryALTinterwordspacing
A.~Caubri{\`{e}}re, N.~A. Tomashenko, A.~Laurent, E.~Morin, N.~Camelin, and
  Y.~Est{\`{e}}ve, ``Curriculum-based transfer learning for an effective
  end-to-end spoken language understanding and domain portability,''
  \emph{CoRR}, vol. abs/1906.07601, 2019. [Online]. Available:
  \url{http://arxiv.org/abs/1906.07601}
\BIBentrySTDinterwordspacing

\bibitem{DBLP:journals/corr/abs-1904-03670}
\BIBentryALTinterwordspacing
L.~Lugosch, M.~Ravanelli, P.~Ignoto, V.~S. Tomar, and Y.~Bengio, ``Speech model
  pre-training for end-to-end spoken language understanding,'' \emph{CoRR},
  vol. abs/1904.03670, 2019. [Online]. Available:
  \url{https://arxiv.org/abs/1904.03670}
\BIBentrySTDinterwordspacing

\bibitem{dinarelli2020data}
M.~Dinarelli, N.~Kapoor, B.~Jabaian, and L.~Besacier, ``A data efficient
  end-to-end spoken language understanding architecture,'' 2020.

\bibitem{DBLP:journals/csl/ManguBS00}
\BIBentryALTinterwordspacing
L.~Mangu, E.~Brill, and A.~Stolcke, ``Finding consensus in speech recognition:
  word error minimization and other applications of confusion networks,''
  \emph{Computer Speech {\&} Language}, vol.~14, no.~4, pp. 373--400, 2000.
  [Online]. Available: \url{https://doi.org/10.1006/csla.2000.0152}
\BIBentrySTDinterwordspacing

\bibitem{zhong-etal-2018-global}
V.~Zhong, C.~Xiong, and R.~Socher, ``Global-locally self-attentive encoder for
  dialogue state tracking,'' in \emph{Proceedings of the 56th Annual Meeting of
  the Association for Computational Linguistics (Volume 1: Long Papers)}.\hskip
  1em plus 0.5em minus 0.4em\relax Melbourne, Australia: Association for
  Computational Linguistics, Jul. 2018, pp. 1458--1467.

\bibitem{DBLP:journals/corr/abs-1810-09587}
\BIBentryALTinterwordspacing
L.~Ren, K.~Xie, L.~Chen, and K.~Yu, ``Towards universal dialogue state
  tracking,'' \emph{CoRR}, vol. abs/1810.09587, 2018. [Online]. Available:
  \url{http://arxiv.org/abs/1810.09587}
\BIBentrySTDinterwordspacing

\bibitem{DBLP:journals/corr/MrksicSWTY16}
\BIBentryALTinterwordspacing
N.~Mrksic, D.~{\'{O}}. S{\'{e}}aghdha, T.~Wen, B.~Thomson, and S.~J. Young,
  ``Neural belief tracker: Data-driven dialogue state tracking,'' \emph{CoRR},
  vol. abs/1606.03777, 2016. [Online]. Available:
  \url{http://arxiv.org/abs/1606.03777}
\BIBentrySTDinterwordspacing

\bibitem{henderson-etal-2014-second}
\BIBentryALTinterwordspacing
M.~Henderson, B.~Thomson, and J.~D. Williams, ``The second dialog state
  tracking challenge,'' in \emph{Proceedings of the 15th Annual Meeting of the
  Special Interest Group on Discourse and Dialogue ({SIGDIAL})}.\hskip 1em plus
  0.5em minus 0.4em\relax Philadelphia, PA, U.S.A.: Association for
  Computational Linguistics, Jun. 2014, pp. 263--272. [Online]. Available:
  \url{https://www.aclweb.org/anthology/W14-4337}
\BIBentrySTDinterwordspacing

\bibitem{Chen2020SchemaGuidedMD}
\BIBentryALTinterwordspacing
L.~Chen, B.~Lv, C.~Wang, S.~Zhu, B.~Tan, and K.~Yu, ``Schema-guided
  multi-domain dialogue state tracking with graph attention neural networks,''
  in \emph{AAAI 2020}, 2020. [Online]. Available:
  \url{https://www.aaai.org/Papers/AAAI/2020GB/AAAI-ChenL.10030.pdf}
\BIBentrySTDinterwordspacing

\bibitem{budzianowski-etal-2018-multiwoz}
\BIBentryALTinterwordspacing
P.~Budzianowski, T.-H. Wen, B.-H. Tseng, I.~Casanueva, S.~Ultes, O.~Ramadan,
  and M.~Ga{\v{s}}i{\'c}, ``{M}ulti{WOZ} - a large-scale multi-domain
  wizard-of-{O}z dataset for task-oriented dialogue modelling,'' in
  \emph{Proceedings of the 2018 Conference on Empirical Methods in Natural
  Language Processing}.\hskip 1em plus 0.5em minus 0.4em\relax Brussels,
  Belgium: Association for Computational Linguistics, Oct.-Nov. 2018, pp.
  5016--5026. [Online]. Available:
  \url{https://www.aclweb.org/anthology/D18-1547}
\BIBentrySTDinterwordspacing

\bibitem{DBLP:journals/corr/abs-1907-01669}
\BIBentryALTinterwordspacing
M.~Eric, R.~Goel, S.~Paul, A.~Sethi, S.~Agarwal, S.~Gao, and
  D.~Hakkani{-}T{\"{u}}r, ``Multiwoz 2.1: Multi-domain dialogue state
  corrections and state tracking baselines,'' \emph{CoRR}, vol. abs/1907.01669,
  2019. [Online]. Available: \url{http://arxiv.org/abs/1907.01669}
\BIBentrySTDinterwordspacing

\bibitem{Ladhak2016LatticeRnnRN}
F.~Ladhak, A.~Gandhe, M.~Dreyer, L.~Mathias, A.~Rastrow, and B.~Hoffmeister,
  ``Latticernn: Recurrent neural networks over lattices,'' in
  \emph{INTERSPEECH}, 2016.

\bibitem{jagfeld-vu-2017-encoding}
G.~Jagfeld and N.~T. Vu, ``Encoding word confusion networks with recurrent
  neural networks for dialog state tracking,'' in \emph{Proceedings of the
  Workshop on Speech-Centric Natural Language Processing}.\hskip 1em plus 0.5em
  minus 0.4em\relax Copenhagen, Denmark: Association for Computational
  Linguistics, Sep. 2017, pp. 10--17.

\bibitem{masamura-et-al-2018}
R.~Masumura, Y.~Ijima, T.~Asami, H.~Masataki, and R.~Higashinaka, ``Neural
  confnet classification: Fully neural network based spoken utterance
  classification using word confusion networks,'' 04 2018, pp. 6039--6043.

\bibitem{Pennington14glove:global}
J.~Pennington, R.~Socher, and C.~D. Manning, ``Glove: Global vectors for word
  representation,'' in \emph{In EMNLP}, 2014.

\bibitem{hashimoto-etal-2017-joint}
\BIBentryALTinterwordspacing
K.~Hashimoto, C.~Xiong, Y.~Tsuruoka, and R.~Socher, ``A joint many-task model:
  Growing a neural network for multiple {NLP} tasks,'' in \emph{Proceedings of
  the 2017 Conference on Empirical Methods in Natural Language
  Processing}.\hskip 1em plus 0.5em minus 0.4em\relax Copenhagen, Denmark:
  Association for Computational Linguistics, Sep. 2017, pp. 1923--1933.
  [Online]. Available: \url{https://www.aclweb.org/anthology/D17-1206}
\BIBentrySTDinterwordspacing

\end{thebibliography}

% \begin{thebibliography}{9}
% \bibitem[1]{Davis80-COP}
%   S.\ B.\ Davis and P.\ Mermelstein,
%   ``Comparison of parametric representation for monosyllabic word recognition in continuously spoken sentences,''
%   \textit{IEEE Transactions on Acoustics, Speech and Signal Processing}, vol.~28, no.~4, pp.~357--366, 1980.
% \bibitem[2]{Rabiner89-ATO}
%   L.\ R.\ Rabiner,
%   ``A tutorial on hidden Markov models and selected applications in speech recognition,''
%   \textit{Proceedings of the IEEE}, vol.~77, no.~2, pp.~257-286, 1989.
% \bibitem[3]{Hastie09-TEO}
%   T.\ Hastie, R.\ Tibshirani, and J.\ Friedman,
%   \textit{The Elements of Statistical Learning -- Data Mining, Inference, and Prediction}.
%   New York: Springer, 2009.
% \bibitem[4]{YourName17-XXX}
%   F.\ Lastname1, F.\ Lastname2, and F.\ Lastname3,
%   ``Title of your INTERSPEECH 2020 publication,''
%   in \textit{Interspeech 2020 -- 20\textsuperscript{th} Annual Conference of the International Speech Communication Association, September 15-19, Graz, Austria, Proceedings, Proceedings}, 2020, pp.~100--104.
% \end{thebibliography}

\end{document}